# "How many images do I need?"

# Understanding how sample size per class affects deep learning model performance metrics for balanced designs in autonomous wildlife monitoring.

Saleh Shahinfar[1], Paul Meek[2, 3], Greg Falzon[1]

[1] School of Science and Technology, University of New England, Armidale, NSW, Australia

[2] NSW Department of Primary Industries, PO Box 530, Coffs Harbour, NSW, Australia

[3] School of Environmental and Rural Science, University of New England, Armidale, NSW, Australia

## Abstract

1. Deep learning (DL) algorithms are the state of the art in automated classification of wildlife camera trap images. The challenge is that the ecologist cannot know in advance how many images per species they need to collect for model training in order to achieve their desired classification accuracy. In fact there is limited empirical evidence in the context of camera trapping to demonstrate that increasing sample size will lead to improved accuracy.

2. In this study we explore in depth the issues of deep learning model performance for progressively increasing per class (species) sample sizes. We also provide ecologists with an approximation formula to estimate how many images per animal species they need for certain accuracy level *a priori*. This will help ecologists for optimal allocation of resources, work and efficient study design.

3. In order to investigate the effect of number of training images; seven training sets with 10, 20, 50, 150, 500, 1000 images per class were designed. Six deep learning architectures namely ResNet-18, ResNet-50, ResNet-152, DnsNet-121, DnsNet-161, and DnsNet-201 were trained and tested on a common exclusive testing set of 250 images per class. The whole experiment was repeated on three similar datasets from Australia, Africa and North America and the results were compared. Simple regression equations for use by practitioners to approximate model performance metrics are provided. Generalizes additive models (GAM) are shown to be effective in modelling DL performance metrics based on the number of training images per class, tuning scheme and dataset.

4. Overall, our trained models classified images with 0.94 accuracy (ACC), 0.73 precision (PRC), 0.72 true positive rate (TPR), and 0.03 false positive rate (FPR). Variation in model performance metrics among datasets, species and deep learning architectures exist and are shown distinctively in the discussion section. The ordinary least squares regression models explained 57%, 54%, 52%, and 34% of expected variation of ACC, PRC, TPR, and FPR according to number of images available for training. Generalized additive models explained 77%, 69%, 70%, and 53% of deviance for ACC, PRC, TPR, and FPR respectively.

5. Predictive models were developed linking number of training images per class, model, dataset and performance metrics. The ordinary least squares regression and Generalised additive models developed provides a practical toolbox to estimate model performance with respect to different numbers of training images.

## 1. Introduction

The rise of camera trapping as a scientific and management data collection tool has forged an unstoppable path across the globe (Meek *et al.*, 2015; O'Connell, Nichols & Karanth, 2011; Rovero *et al.*, 2013; Swann & Perkins, 2014). The consequence has created a hiatus between data capture and data storage/analysis because of the large volumes of image data collected and the manual analysis of that data (Falzon *et al.*, 2014; Meek *et al.*, 2019; Price-Tack *et al.*2016; Scotson *et al.*, 2017). Automating the analysis of camera trap data has been recognised as a critical step forward (Falzon *et al.*, 2020; Meek *et al.*, 2015; Nazir *et al.*, 2017) although there are enormous difficulties, challenges and technological impediments to creating robust automated analytical tools (Falzon *et al.*, 2020; Meek *et al.*, 2019). This pressing need to develop reliable automated software to automatically scan, detect and sort data for further analysis has led to the age of big data and the subsequent integration of field ecology and ecological informatics.

Arguably one of the most promising avenues towards integrating computer assisted technologies into camera trapping has been the automated detection of animals in image and video data through Deep Learning algorithms. Convolutional neural networks (CNN) in particular have been used in a range of ecological studies including the classification of animal species in camera trap images (Gomez Villa *et al.*, 2017; Norouzzadeh *et al.*, 2018; Tabak *et al.*, 2019); detecting mammals via unmanned aerial vehicles (Kellenberger *et al.*, 2018) and robust detection and re-identification of marine animals (Moskvyak *et al.*, 2019; Xu & Matzner, 2018) are increasingly finding further applications in the monitoring of wildlife and livestock.

Practical application of CNN algorithms can be challenging and meets with a range of difficulties. A major challenge is that camera trap studies often produce highly unbalanced datasets with a small number of image classes (e.g. birds and common species) comprising the vast amounts of images and other classes (e.g. Koalas, or endangered species) containing far less imagery. This issue of class imbalance has potential to cause major bias in the models developed and recent research efforts have explored the best approaches for ecological studies (Clare *et al.*, 2019; Kellenberger *et al.*, 2018). A promising avenue forward is the generation of synthetic images to improve classifier performance for the recognition of rare classes (Beery et al. 2020). Another major issue is the high degree of visual similarity in camera trap image datasets. Thousands of images can be collected from the same site which can lead to *over-learning* of the CNN model and instability when applied to new study sites (Beery, Van Horn & Perona 2018). Falzon et al. (2020) resolved this issue by developing optimised models for each study site. Such approaches are particularly suitable for large long-term monitoring projects where resources can be invested in the development of top-performing models. These models are inherently valuable for major ecological management projects but are distinctly different from general purpose models which can be used across the camera trapping community. This research will focus on the scenario where models are being developed for larger-scale longer-term monitoring programs. In such scenarios, a region is monitored and sufficient camera trap imagery collected before commencing model training. This project-specific model is then verified for performance and integrated into the research program to assist with wildlife monitoring.

A key question faced by practitioners developing CNN-based species recognition algorithms is "*how many images per class are required to achieve a certain level of accuracy?*". In other words what is the relationship between number of images of a species in the training set and the expected model performance metrics (MPM). This information would be extremely useful for researchers to know when planning data collection exercises in order to develop the species recognition models. A survey of the literature provides barely any information pertaining to camera trapping on this critical topic. Logarithmic trends between model performance and training data sample size have been reported on various computer vision task utilising transfer learning on JFT-300M and COCO datasets (Sun *et al.*, 2017). In the medical imaging field Cho *et al.* (2016) provided an inverse power law function for predicting the expected accuracy for the number of images available per class for training. Tabak *et al.* (2019) briefly report a logarithmic trend curve when exploring the performance of ResNet-18 models on a North American data set. As CNN models become more widely adopted by the camera trap community there is a pressing need to gain a better understanding of this issue in order to both guide model developers and also to ensure quality control of the models developed. The relation between MPM such as accuracy with the number of images per class, the variation of MPM and the differences encountered across field sites, different CNN architectures and depths along with different transfer learning strategies all need to be explored. Finding a predictable relationship between MPM and number of images per class would also be highly useful and guide practitioners. The objectives of this research are to explore these issues in the context of *project-specific models* and to provide useful guidelines when developing CNN for species recognition.

## 2. Materials and Methods

## 2.1. Data

Three distinct datasets from Australia, Africa and North America were used in this research in order to provide geographically and taxonomically diverse locations.

### 2.1.1. Snapshot Australia

This dataset was collected from camera traps located along forestry trails along the Great Dividing Range in the state of New South Wales, Australia. Reconyx HC600 camera traps were set to "burst" mode to capture ten images per trigger. In total this dataset included 29 distinct species of native Australian animals and livestock.

### 2.1.2. Snapshot Serengeti

The Snapshot Serengeti project has been the largest publicly available species level annotated camera-trap image dataset in the world consisting of 7 million images of which 1.2 million images contained animals. The data set contains images of 54 African wildlife species collected throughout the Serengeti National Park in Tanzania (Swanson *et al.*, 2015; Willi *et al.*, 2019).

### 2.1.3. Snapshot Wisconsin

The Snapshot Wisconsin project is a network of cameras located across the state of Wisconsin, USA. It is a joint government, university, industry and public collaboration which provides year-round monitoring across all counties using camera traps. In total this dataset included 43 distinct species of animals (Willi *et al.*, 2019; Wisconsin Department of Natural Resources, 2019).

The source data from the Australian, Serengeti and Wisconsin projects were randomly sub-sampled for analysis and comparisons. Each data set sample includes 8 classes, consisting of 6 animal species classes, one "empty" image class which consists of only vegetation with no animal being present, and one category called "others" which is formed by a random combination of camera trap images of other animal species not allocated to their own specific class. Selection of the 6 animal species classes for each data set was determined by selecting classes which occurred with high frequency and therefore represent a large component of the workload for human annotators. Rare classes with insufficient numbers were not included in this study and were instead aggregated into the 'Other' class which requires review by a human post-processing. A total of 6 species classes was set across datasets to facilitate comparisons noting that some datasets such as Snapshot Serengeti might have been able to accommodate more species classes in practice. A balanced design was generated by ensuring an equal number of samples was generated for each class. Balanced designs were selected to facilitate model comparisons and to reduce model bias to particular classes.

A high-degree of spatio-temporal correlation was observed in the datasets as many images were captured by each camera and there were often multiple triggers of the camera as an animal walked past. Figure 1 (a)-(c) displays an example of a highly spatio-temporally correlated image sequence which was captured as a feral cat walked past a camera trap. The concern with highly correlated datasets is the high level of visual similarity between images. High levels of similarity could reduce the capacity of the CNN algorithms to learn robust distinctive image features and it might not provide realistic levels of variation in the appearance of animals in the images of the test data set. In order to address these concerns, sub-sampling was used. From the original dataset 750 of each class selected with equal spacing across the entire time-date ordered dataset to avoid similarity of adjacent images as much as possible. One third of these images (250 images) selected randomly and was excluded from the dataset as an exclusive "test set". From the remaining images, 6 training sub-sets were drawn at random with 10, 20, 50, 150, 500, and 1000 images each. In order to safeguard against bias of classes to particular locations we reviewed both the source training and test image datasets to confirm that there were 3 locations or more present for each class. The source datasets used generally span years, seasons and environmental conditions in order to form a representative sample of the region. Table A1 in Appendix A, provides the class composition details for each dataset.

## 2.2. Deep Learning Models

Deep neural networks are successors of feedforward networks with four distinct properties: more neurons and layers, more complex connections between layers, automatic feature selection and need of high computation power to train (Patterson & Gibson, 2017). As a form of representation learning, deep learning model is capable of receiving raw data and automatically discovering the representation needed for pattern recognition. Multiple levels of representations obtained by composing simple non-linear transforms of the input data in order to learn a complex function (LeCun, Bengio & Hinton, 2015).

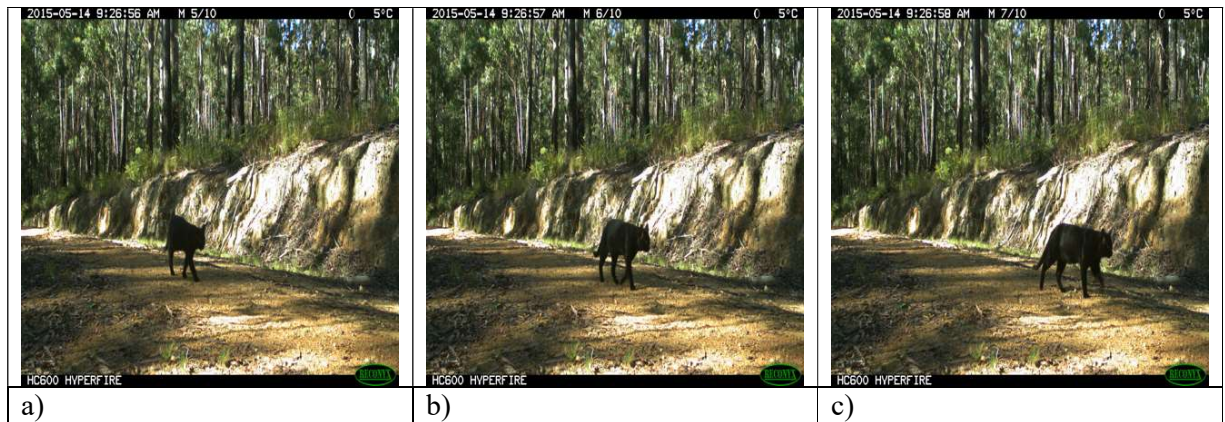

| a) | b) | c) |

Figure 1: Spatio-temporal correlation in camera trap image sequence data.

### 2.2.1 Deep Learning Architectures

Deep learning networks are distinguished from each other by their architectures in the form of the number of convolutional layers and neurons, type of connection between layers, activation functions, and optimization algorithms. In this study, two families of deep learning were used; Residual Neural Networks (ResNet) and Densely Connected Neural Networks (DenseNet). From each family we used three variants; resNet-18, resNet-50, and resNet-152 of ResNet family, and dnsNet-121, dnsNet-161, and dnsNet-201 were used from DenseNet family. The ResNet and DenseNet CNN families were selected primarily due to their capability to increase the number of layers and in the case of ResNet its previous demonstration as being competitive option for automated species recognition (Norouzzadeh et al. 2018; Tabak et al 2019; Willi et al 2019). Another good candidate architecture was VGG family of networks due to their high performance in Norouzzadeh et al., (2018) however they were excluded from the candidate networks due to their comparatively lengthy training process.

#### 2.2.1.1. Residual Neural Networks

Deep residual networks faces the problem of vanishing gradients whereby model accuracy plateaus and then decreases sharply as network depth increases (Bengio *et. al*., 1994; He *et al*., 2016). ResNet architectures overcome this issue by adding short-cut connections between each 2-3 consecutive convolutional layers in the network. This subtle but effective change solves the problem of vanishing gradients by adding activation from a few layers behind at no extra parameter cost and also decreases the training time. To date ResNet is one of the most efficient and accurate DL architectures achieving first place results in both the Imagenet Large Scale Visual Recognition Challenge 2015 and Common Objects in Context 2015 competitions.

#### 2.2.1.2. Densely Connected Neural Networks

DenseNet extends on the short-cut connections introduced in ResNet by connecting every layer to every subsequent layer in feed-forward form (Huang *et al.,* 2017). Each layer uses the feature map of all previous layers as input. This architecture overcomes the vanishing gradients problem, improves feature propagation, facilitates feature reuse, and reduces the number of model parameters significantly (Huang *et al.*, 2017). The DenseNet architecture has been found to provide similar accuracy levels to ResNet but with higher model complexity which might be a limitation for embedded vision devices such as smart cameras but also potentially allow greater learning capacity for the model across a larger number of classes or environments.

## 2.2.2. Deep Learning Training

A key step in DL is model training whereby network weights are learned through the application of labelled data and stochastic optimisation algorithms. Application of the model training process from "scratch" requires large datasets with millions of labelled images along with high-performance hardware. Assembly of such datasets requires extensive effort whilst the access to hardware along with the technical knowledge required can make this approach prohibitive. Fortunately practitioners can utilise a machine learning technique called *transfer learning* to modify a model trained to do a similar task in order to achieve a target task of a study. Transfer learning uses the knowledge generated on one task to achieve a different but similar in nature task (Norouzzadeh *et al.*, 2018; Yosinski *et al.*, 2014). For example a CNN used to classify flowers could be modified using transfer learning and to produce a CNN which classifies fish. One of the primary benefits of transfer learning is that far less labelled data is often required to develop the new CNN for a target task. The computational complexities are also often greatly reduced. This is a result of the older source CNN generating knowledge about the data features which are often share a degree of similarity with the target domain data set. As a result of transfer learning the data annotation and computational costs are greatly reduced which makes it feasible to utilise Deep Transfer Learning approaches in the context of camera trap images. In this study, all models were initialised from a source CNN structure derived from the ImageNet data set which consists of 14,197,122 labelled images of a wide range of classes including animals, birds, fish, flowers, humans, tools, trees and vehicles among others (Deng *et al.*, 2009). The ImageNet dataset is widely used in the computer vision community to produce source network weights in order to perform transfer learning on image datasets.

### 2.2.2.1. Training Scheme

Model training and testing was conducted in fastai v01 framework; a PyTorch based library (Howard, 2018; Paszke *et al.*, 2017). Each network was trained twice per dataset, once with shallow tuning and once with deep tuning with 25 and 50 epochs respectively. Shallow tuning refers to the task which initialises the network with pre-trained weights and modifies only the last fully connected layers when re-training with the new training set (Yosinski *et al.*, 2014). In this technique usually pre-trained network based on ImageNet are used. Deep tuning on the other hand is when all the layers (or at least several layers) of the pre-trained network's weights are allowed to be modified when training the network for a new dataset (Tajbakhsh *et al.*, 2016). All networks were trained using Adaptive Moment estimation (ADAM) backpropagation algorithm with momentum and weight decay to optimise the CrossEntropy loss. Maximum learning rate was set to $\lambda$ = 1e-03 for all the networks and the best learning rate was derived via cyclical learning rate policy (Smith, 2018) by internal cross validation on a 30% subsample of training set. A batch size of bs = 8 was used for all the trainings, as this batch size is amenable for training across a range of Graphical Processing Unit hardware.

Data augmentation was used to supplement model training data and artificially introduced random variations in order to enhance the robustness of the model developed. The data augmentation technique consists of applying small random stochastic transformations to the training image data set and using both the source training data along with the augmented data in the model training process. Use of data augmentation has been found to improve model robustness and increase test accuracy (Krizhevsky, Sutskever & Hinton, 2012; Zoph *et al.*, 2019). Seven standard data augmentation practices were applied consisting random cropping, horizontal flipping, rotation, zooming in, symmetric wrapping, and changing brightness and contrast of the images (Krizhevsky, Sutskever & Hinton 2012; Zoph *et al.*, 2019). Data augmentation can also be applied on the test dataset (Howard, 2018). The use of data augmentation increased the training set by 8-fold (seven types of augmentation and the original data set).

In addition, four different data augmentation schemes were implemented to generate pseudo-replicates for model development purposes. These augmentation strategies were:

1- Training set augmentation; No testing set augmentation
2- Training set augmentation; Testing set augmentation
3- No training set augmentation; Testing set augmentation
4- No training set augmentation; No testing set augmentation

These pseudo-repetitions were used in regression model development later in this study.

Model training and testing was performed on a Lambda Quad machine with 4 GPUs (RTX 2080 Ti); 12 dual core CPUs (Intel i9-7920X; 2.90 GHz), and 128 gigabytes of RAM. For each tuning, augmentation, dataset, architecture, and training size combination the model with lowest validation loss was selected as the best model and used for prediction on the testing set.

### 2.2.3. Model Performance Metrics

Four key metrics were evaluated based on the need to critically examine different aspects of model performance.

Per class Accuracy (ACC) provides an overall measure of performance unweighted to the type of classification error which occurred, e.g. what proportion of "kangaroo" images were correctly classified by the model? The potential drawback of accuracy is that it is not a good measure of performance when there is severe class imbalance, however it is not the case in our study as we have balanced study design.

$$ACC = \frac{TP + TN}{TP + FP + TN + FN} \quad (1)$$

Where the symbols indicate the number of $TP$ = true positives, $FP$ = false positives, $TN$ = true negatives, and FN = false negatives.

Per class Precision (PRC) describes the fraction of actual correctly classified images to the total number of images classified by the model as that class, e.g. what proportion of images classified as "kangaroos" were actually labelled as containing kangaroos? Precision is a useful measurement when the cost of FPs are high in which we want avoiding FP classification by our model as much as possible.

$$PRC = \frac{TP}{TP + FP} \quad (2)$$

Per class True Positive Rate (TPR) otherwise known as *recall* describes the proportion of correctly classified images per class, e.g. what proportion of "kangaroo" images were correctly classified? Recall is important when the cost of FNs are high and need to be avoided.

$$TPR = \frac{TP}{TP + FN} \quad (3)$$

Per class False Positive Rate (FPR), describes the fraction of incorrectly classified images from all negative images, e.g. what proportion of "non-kangaroo" images are classified incorrectly as "kangaroo"? Similar to precision, FPR are useful when the model is being evaluated for its ability to avoid FP predictions.

$$FPR = \frac{FP}{FP + TN} \quad (4)$$

## 2.3 Predictive Modelling

Ordinary least squares regression models were used to guide practitioners on the expected MPM for a particular training sample size. In addition generalised additive models (GAM) (Hastie & Tibshirani, 1986; Hastie & Tibshirani, 1990) were utilised to provide a detailed understanding of the significant factors governing model performance. The GAM is an extension of the generalised linear model which includes a linear predictor term which is the sum of smooth functions of covariates (Wood, 2017). The GAM has the following form:

$$g(\mu_i) = X_i^* \theta + \sum_j f_j(x_{ij}) + \varepsilon \quad (5)$$

$$\varepsilon \sim N(0, \sigma^2)$$

Where $\mu_i \equiv E(Y_i)$ is the expected value of the response variable and $Y_i$ is the response variable with an assumed exponential family distribution. $X_i^*$ is a row of incidence matrix for any parametric components of the model, $\theta$ is the corresponding parameter vector, $f_j$ is the jth smooth function of $x_{ij}$ covariates and $\varepsilon$ is the additive error term which is assumed to follow a Normal distribution. Smooth functions often are built via reduced rank smoothing approach due to lower computational cost than other methods (Wood, 2011, 2017). Exploratory analysis indicated that the $Y_i$ distributions of MPM were found to resemble the Beta distribution (Figure 2), therefore a GAM with a Beta regression family and '*logit*' link function was used for model fitting and evaluation.

Predictor variables for MPMs (ACC, PRC, TPR, and FPR) were: 1) model tuning; specifically the type of tuning considered on the DL architecture with two levels, shallow and deep. 2) dataset; with three levels indicating the study location, namely; Australia (AU), Serengeti (SE) and Wisconsin(WI). 3) six deep learning architectures (dlArch) including: resNet-18, resNet-50, resNet-152, dnsNet-121, dnsNet-161, dnsNet-201. 4) number of images in the training set (numTrImages) with six levels; 10, 20, 50, 150, 500, 1000 images per class.

A GAM was fitted to each MPM of interest separately using the R package mgcv (Wood, Pya, N., & Saefken, B. (2016). Each MPM measurement for each category of image were our observation units in this analysis which modelled as $y_i \sim Beta(\alpha, \beta)$ where $\alpha$ and $\beta$ are shape parameters where $\frac{\alpha}{\alpha+\beta}$ and $\frac{\alpha\beta}{(\alpha+\beta)^2 * (\alpha+\beta+1)}$ define mean and variance of the distribution respectively. Statistical models considered in this study derived via backward stepwise elimination on the full model including two-way interaction terms and smoothing functions for all the predictors. To eliminate any terms in the model a 5% level of significance was considered and any main effect or interaction that did not meet the 5% significant level was excluded from the model. Backward elimination approach was chosen because we assumed that all the measured network characteristics such as tuning and augmentation might have effect on the model performance metrics. The final model for each performance measurements follows:

$$y_i = t_{ji} + d_{ki} + a_{li} + \sum_{m=1}^{5} b_{nm}(n_i)\beta_{nm[d_{ki}]} + \varepsilon \quad (6)$$

$$\varepsilon \sim N(0, \sigma^2)$$

Where $t_{ji}$ is the fixed effect of tuning scheme with two levels except for FPR, (in backward elimination, tuning scheme was found insignificant for prediction of FPR and hence were removed from the predictor variables), $d_{ki}$ is the fixed effect of data set with three levels, $a_{li}$ is the fixed effect of DL architecture with six levels, $b_{mk}$ is the reduced ranking cubic spline basis function for number of training images i.e. $n_i$, nested in data set $k$; and $\beta_{nm[d_{ki}]}$ is the regression coefficient for the numTrImages nested in data set $d_{ki}$ with $m = 5$ knots being placed between each numTrImage magnitude.

Following the GAM analysis, linear regression counterparts of the GAM models were fitted according to the following form:

$$y \sim \log(numTrImages) + \varepsilon \quad (7)$$

$$\varepsilon \sim N(0, \sigma^2)$$

Where $y$ corresponds to the ACC, PRC or TPR metric of interest. When $y$ was corresponds to FPR, the following model was fitted instead.

$$y \sim \log(1/numTrImages) + \varepsilon \quad (8)$$

$$\varepsilon \sim N(0, \sigma^2)$$

## 3. Results

Model performance metrics across all datasets, models and training schemes are first presented followed by the results of the predictive models.

### 3.1. Model Performance Metrics

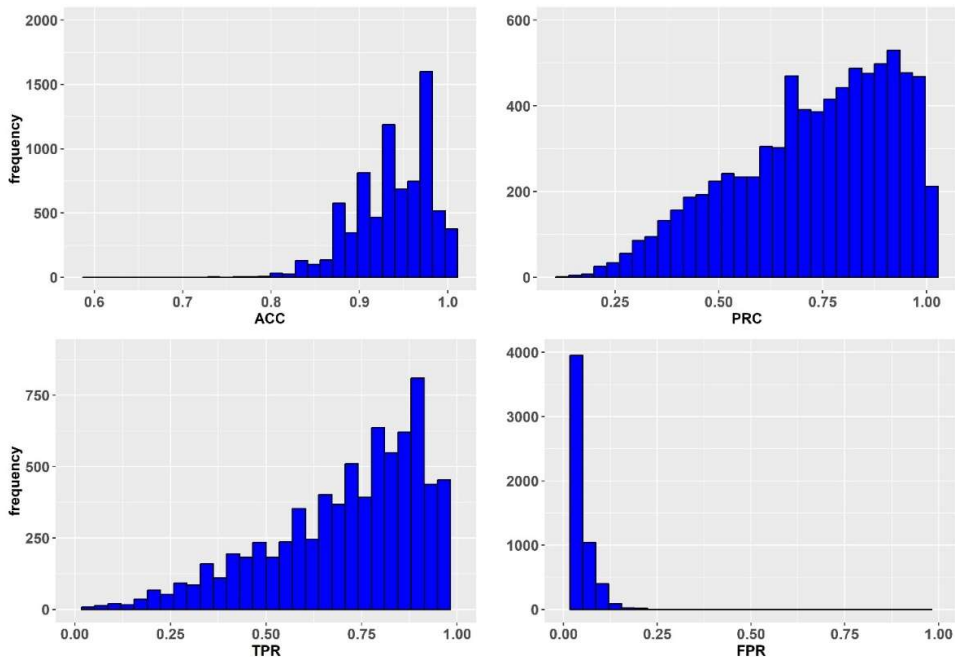

Figure 2. Histogram of model performance metrics aggregated for all the models and datasets. In total 864 model comprising [3 data sets; 6 training set sizes; 6 deep network architecture; 4

augmentation scenarios; 2 tuning schedules] (Observe the significant variation and high skew. ACC: accuracy, PRC: precision, TPR: True Positive Rate and FPR: False Positive Rate.)

### 3.2 Predictive Modelling

Table 1. Ordinary Least Square Regression Models and Parameters to predict model performance metrics a priori based on number of available images in training set (for balanced design).

| Model Performance Measurement | Intercept | Slope | Adj-R^2 |
|---|---|---|---|
| ACC | 0.85 | 0.02*log(numTrImages) | 0.57 |
| PRC | 0.34 | 0.09*log(numTrImages) | 0.54 |
| TPR | 0.32 | 0.09*log(numTrImages) | 0.52 |
| FPR | 0.09 | 0.01*log(1/numTrImages) | 0.34 |

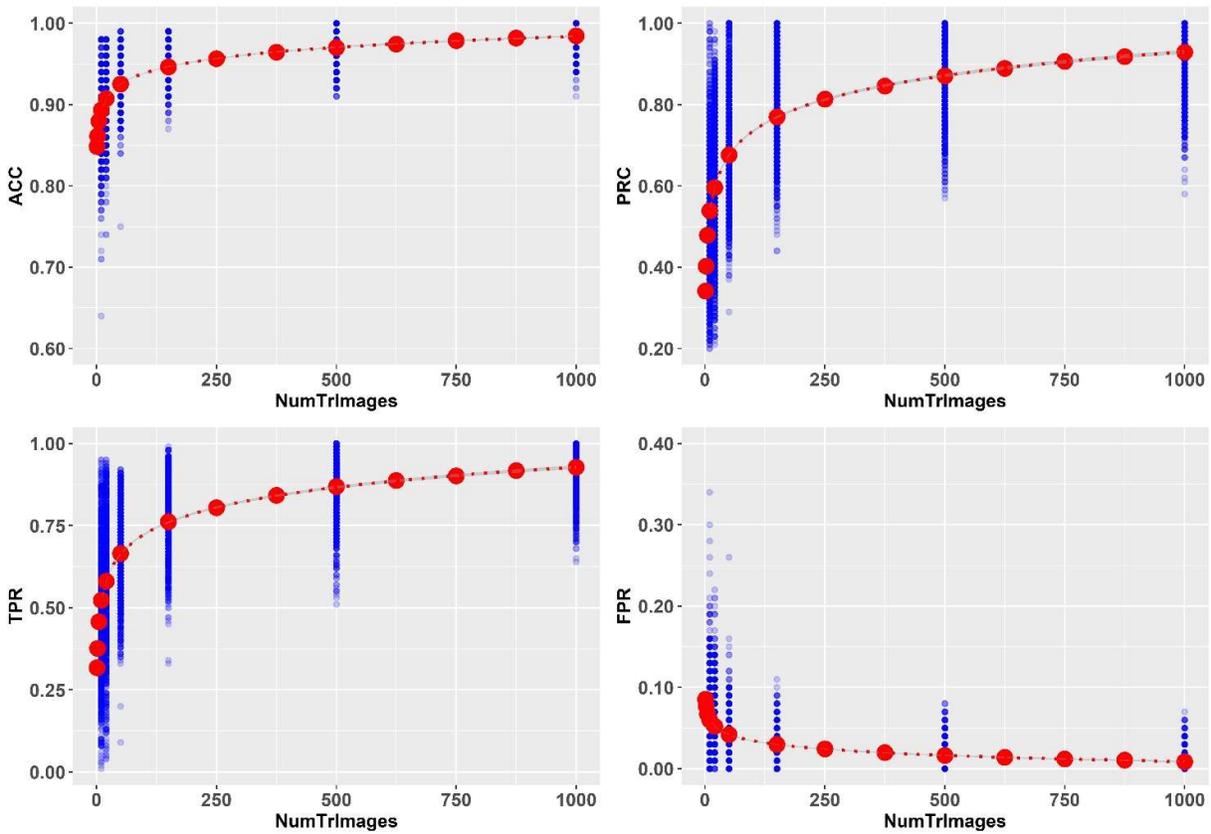

Figure 3. Model Performance Metrics scale according to the number of training images (NumTrImages). Blue dots indicate realised model performance metrics irrespective of dataset, model architecture, tuning and augmentation strategy. Red lines and dots indicate predicted model performance metrics according to Ordinary Least Square Estimates.

Table 2. Generalised Additive Model summaries for prediction of accuracy (ACC) and precision (PRC).

|  | ACC | | | PRC | | |
|---|---|---|---|---|---|---|
| **Parameters** | **Estimates** | **Std.Error** | **P-Value** | **Estimates** | **Std.Error** | **P-Value** |
| (Intercept) | 3.322 | 0.016 | 2.00e-16 | 1.591 | 0.023 | 2.00e-16 |
| tuningShallow | 0.050 | 0.009 | 1.23e-07 | 0.045 | 0.015 | 2.68e-3 |
| dataSetSE | -0.452 | 0.014 | 2.00e-16 | -0.414 | 0.020 | 2.00e-16 |
| dataSetWI | -0.747 | 0.014 | 2.00e-16 | -0.845 | 0.019 | 2.00e-16 |
| dlArchdnsNet161 | 0.068 | 0.017 | 4.12e-05 | 0.095 | 0.026 | 2.363e-4 |
| dlArchdnsNet201 | 0.044 | 0.017 | 0.008.67e-3 | 0.064 | 0.026 | 1.34e-2 |
| dlArchresNet152 | -0.053 | 0.016 | 0.001.17e-3 | -0.072 | 0.026 | 5.19e-3 |
| dlArchresNet18 | -0.123 | 0.016 | 2.65e-14 | -0.178 | 0.026 | 2.77e-12 |
| dlArchresNet50 | -0.059 | 0.016 | 2.76e-4 | -0.077 | 0.026 | 2.57e-3 |
| *Smoothing Functions* | **EDF** | **Ref.df** | **P-Value** | **EDF** | **Ref.df** | **P-Value** |
| *s(numTrImages):dataSetAU* | 3.901 | 4 | <2e-16 | 3.952 | 4 | <2e-16 |
| *s(numTrImages):dataSetSE* | 3.968 | 4 | <2e-16 | 3.966 | 4 | <2e-16 |
| *s(numTrImages):dataSetWI* | 3.944 | 4 | <2e-16 | 3.957 | 4 | <2e-16 |
| *R-sq.(adj)* | 0.660 | | | 0.630 | | |
| *Deviance explained* | 0.773 | | | 0.689 | | |

Table 3. Generalised Additive Model summaries for prediction of True Positive Rate (TPR) and False Positive Rate (FPR).

|  | TPR | | | FPR | | |
|---|---|---|---|---|---|---|
| **Parameters** | **Estimate** | **SE** | **P-Value** | **Estimate** | **SE** | **P-Value** |
| (Intercept) | 1.460 | 0.023 | 2.00e-16 | -3.797 | 0.021 | 2.00e-16 |
| tuningShallow | 0.074 | 0.015 | 4.30e-07 | - | - | - |
| dataSetSE | -0.388 | 0.019 | 2.00e-16 | 0.333 | 0.019 | 2.00e-16 |
| dataSetWI | -0.795 | 0.019 | 2.00e-16 | 0.628 | 0.018 | 2.00e-16 |
| dlArchdnsNet161 | 0.091 | 0.026 | 4.08e-4 | -0.078 | 0.025 | 0.001.55e-3 |
| dlArchdnsNet201 | 0.063 | 0.026 | 1.39e-2 | -0.045 | 0.024 | 0.06.45e-2 |
| dlArchresNet152 | -0.083 | 0.025 | 9.93e-4 | 0.033 | 0.024 | 1.67e-1 |
| dlArchresNet18 | -0.171 | 0.025 | 1.12e-11 | 0.126 | 0.024 | 1.05e-07 |
| dlArchresNet50 | -0.093 | 0.025 | 2.44e-4 | 0.025 | 0.024 | 3.02e-1 |
| **Smoothing Functions** | **EDF** | **Ref.df** | **P-Value** | **EDF** | **Ref.df** | **P-Value** |
| s(numTrImages):dataSetAU | 3.922 | 4 | <2e-16 | 3.821 | 4 | <2e-16 |
| s(numTrImages):dataSetSE | 3.964 | 4 | <2e-16 | 3.913 | 4 | <2e-16 |
| s(numTrImages):dataSetWI | 3.962 | 4 | <2e-16 | 3.735 | 4 | <2e-16 |
| **R-sq.(adj)** | 0.611 | | | 0.390 | | |
| **Deviance explained** | 0.695 | | | 0.531 | | |

## 4. Discussion

### 4.1. Model Performance Metrics

Our findings showing the relationship between the number of training images and accuracy measures of DL networks showed great promise. Aggregated MPM on tuning, augmentation and DL architectures are presented in Tables A2 to A4 in Appendix A. There was no systematic difference in model performance metrics between different categories within each dataset used in this study.

In general, accuracy, precision, TPR, and FPR improves with increasing numbers of images in a training set as we predicted. The relationship between numTrImages and performance metrics shows a logarithmic trend (Figures A1-A4 in Appendix A). This trend is in line with the previous findings in this field (Sun *et al.*, 2017; Tabak *et al.*, 2019). However, in Sun et al (2017) only the mean average precision was reported and in Tabak et al. (2019), only resNet-18 was used as the learning algorithm. In addition Tabak et al. (2019) utilised imbalanced datasets with varying numbers of images per class both in the training and testing datasets. In this study we attempted to address this issue by ensuring that all calculations were performed on multiple balanced training sets. Ensuring balanced datasets removes potential biases in the sample size estimates and informs model performance in ideal scenarios where it is possible to control sampling during model development. The scenario of balanced training data is typically artificial for camera trap datasets and the practitioner is faced with the scenario of either using all available data or alternatively only training the model to recognise categories with sufficient numbers of samples to ensure quality control over any potential biases towards particular categories. To this end the ideal model training strategy is still under exploration and there are potential advantages and disadvantages to each of the aforementioned approaches. Our research provides a better contextual understanding of model performance across environments and the factors that might influence this performance.

Accuracy metrics were assessed for "in-bag" samples - that is samples from the same set of sites and monitoring cameras. This metric evaluation approach is appropriate for the development of classifiers intended for use in long-term monitoring projects. However as in this study we did not address the issue of "model drift", these results cannot guarantee a generalized accuracy many years into the future. In practice, model performance needs to be monitored and where necessary periodically updated with new images if conditions change and model performance begins to decline. Evaluation of performance for "out-of-bag" samples such as different camera models, sites and projects was not conducted. Algorithms to address such scenarios are still under investigation from researchers and routine application of available models in such scenarios is not recommended as currently available approaches can produce unpredictable and unreliable results. We intend to explore such issues in future research, the current investigation concerns the pressing need to advance automated recognition algorithms for larger-scale longer-term studies which can invest the resources to adequately develop data samples and recognition models for their projects

### 4.2. Effect of increasing training images

Overall, a logarithmic trend was observed between ACC, PRC, and TPR and the number of images in the training set. Similarly, FPR was associated with the logarithm of inverse of numTrImages (Figures A1-A4 in Appendix A). ACC increased by about 4-14% as numTrImages increased from 10 to 1000. In contrast, the magnitude of PRC and TPR increased by 15-63% and 12-66% over the same range for PRC and TPR respectively. In terms of FPR 50-83% reduction in magnitude observed as numTrImages increased from 10 to 1000. Typically, after 150 images per class improvement in the performance slowed down with no substantial improvement after 500 images. Therefore we conduced that in case of limited resources, 150-500 images per class is sufficient to achieve reasonable classification accuracy for project-specific camera trap models. These results are in line with (Barz & Denzler, 2020) despite their dataset limitation of only 250 images per class in resNet-110.

### 4.3. Effect of dataset

We report that when numTrImages was 10, 20 or 50 the Serengeti data set had the best performance with Australian data set being second with negligible differences (Tables A2-A4 of Appendix A). However, in 150, 500, and 1000 numTrImages their rank flipped over and Australian dataset shows the best performance and this gap increased as numTrImages increased. This observed difference could be a real effect or due to random sampling effects and it wasn't investigated further. On the other hand Wisconsin dataset had the poorest performance in all cases irrespective of numTrImages. Inferior performance of Wisconsin dataset might be due to the nature of images in this dataset. Most of the data in this dataset are captured at night with object animals being too small and hard to identify even by human eyes therefore poorer performance of this dataset is expected. No systematic variation among performance of DL algorithms on any particular animal class or blank field was observed.

### 4.4. Effect of network architecture

Overall within each network architecture performance improved with increasing numTrImages. Of the two architectures, DenseNet had slightly better performance than ResNet. This trend was more noticeable in case of PRC and TPR than ACC and FPR. Variance of any MPM also decreased 2 to 4 times as numTrImages increased from 10 to 1000. This decrease was negligible when deeper networks were used. In general the network architectures DenseNet-161 and ResNet-18 were associated with highest and lowest performance, respectively, while other architectures performed somewhere in between (Table A5 in Appendix).

### 4.5. Effect of transfer learning scheme

In general, shallow tuning – tuning of only the last network layer – was demonstrated to be an adequate and competitive option for wildlife classification, although the difference was not significant compared to deep tuning. This observation corresponds with Sun et al. (2017) who compared fine (shallow) tuning of a pre-trained resNet-101 on JFT-300M dataset (Hinton, Vinyals & Dean 2014) with random initialization of network weights and training from scratch. Inferior performance of deep tuning might be due to insufficient number of epochs (which was set to 50 epochs). Another reason for that might be existence of the same or similar species of the animals within the ImageNet dataset. Use of pre-trained networks with ImageNet might bias models to more effectively learning particular classes than others. Yosinski *et al.*, (2014) showed that transferability of features decreases as difference between training dataset and target dataset increases. In cases in which training data is limited however, transferring features from distant tasks might be more beneficial than using random weights for initializing the network. Epoch length was limited and loss monitored due to computational considerations and to avoid the risk of drastically over-fitting the model to the training dataset.

### 4.6. GAM analysis

Four generalised additive models were fit to explore and understand the link between model performance metrics and predictor variables. Results of GAM analysis presented in Tables 2 and 3 indicated that model performance metrics considered in this study, were significantly associated with tuning scheme (except for FPR), data set, DL architectures, and numTrImages nested in dataset

### 4.7 Practical Guide for practitioners

The simple linear regression models (Table 1) relates MPM with a logarithmic trend to the numTrImages (Figure 3). These simple linear regression models do not fit the observed data as well as the GAM's (Table 2 & 3) but have the great benefit of providing a general and easy to use guide for estimation of model performance *a priori*. Overall, for balanced studies at long-term monitoring sites the ACC, PRC and TPR metrics scale logarithmically with the number of training images whilst the FPR scales inverse logarithmically. Ideally, the images used in the model training scheme should be as visually distinctive as possible and not repetitions from a camera trap burst sequence. Practitioners should also implement a shallow fine-tuning strategy and use deeper networks were possible.

## 5. Conclusions and Future Works

Artificial intelligence and machine learning holds much promise in the field of camera trapping. Impressive advances have been made in the field of automated recognition of animal species in camera trap images. Such advances have potential to greatly reduce workload of staff and also allow implementation of camera trap monitoring programs at a scale previously unattainable. Key to success of such programs is the development of reliable methodology with a high standard of quality control. Deriving a general relationship between model performance metrics and the number of images used in model training is a key component in the development of a quality assurance process. Sourcing a sufficient number of labelled images for each species category is a key challenge in the practical implementation of deep learning algorithms in the field of camera trapping. Knowing when a sufficient number of images has been collected to ensure reliable identification of particular animal species is of significant practical interest. Therefore in this study we have explored the impact of number of training images, training schemes and network architectures across three different and geographically diverse datasets. Shallow transfer learning and deeper network architectures were found to optimise classification performance. Overall accuracy, precision and recall were identified to have a logarithmic relationship with the number of images in the training set whilst false positive rate was associated with the logarithm of the inverse of the number of images in the training set. This logarithmic relationship held across datasets and is consistent with the findings reported by Tabak et al., (2019), albeit this current study explored the issue in far greater detail and in a controlled balanced sample setting. Importantly, the results reported are most relevant for long-term monitoring sites with a high degree of visual similarity in the image background. Such long-term monitoring projects form an important component of wildlife monitoring however this does not account for other forms of camera trap monitoring projects which survey across a wide range of random sites. In such circumstances the robustness of camera trap recognition algorithms can fail ( Beery, Van Horn & Perona 2018) and there is a need to better understand both the data sampling requirements and factors which can lead to such failures. Similarly the impacts of incorrect labelling, imbalanced samples, image quality, camera trap differences and compositions of different recognition classes needs to be better understood. There is clearly a large amount of further research to be performed in order to improve quality assurances of camera trap image recognition algorithms and also to establish better guidelines to sampling requirements. Nonetheless, this research provides detailed information and guidelines from which to build more reliable automated species recognition algorithms.

## Acknowledgments

We thank Joshua Stover, Norman Gaywood, David Paul, Mitch Welch from University of New England, and Stas Bekman, Solyvan Guagger from fast.ai support team for their kind and helpful supports.
Funding for this project was provided by the Australian Government Department of Agriculture and Water Resources through the *e-Technology Hub* – Utilising Technology to Improve Pest Management Effectiveness and Enhance Welfare Outcomes project.

**Appendix A:**

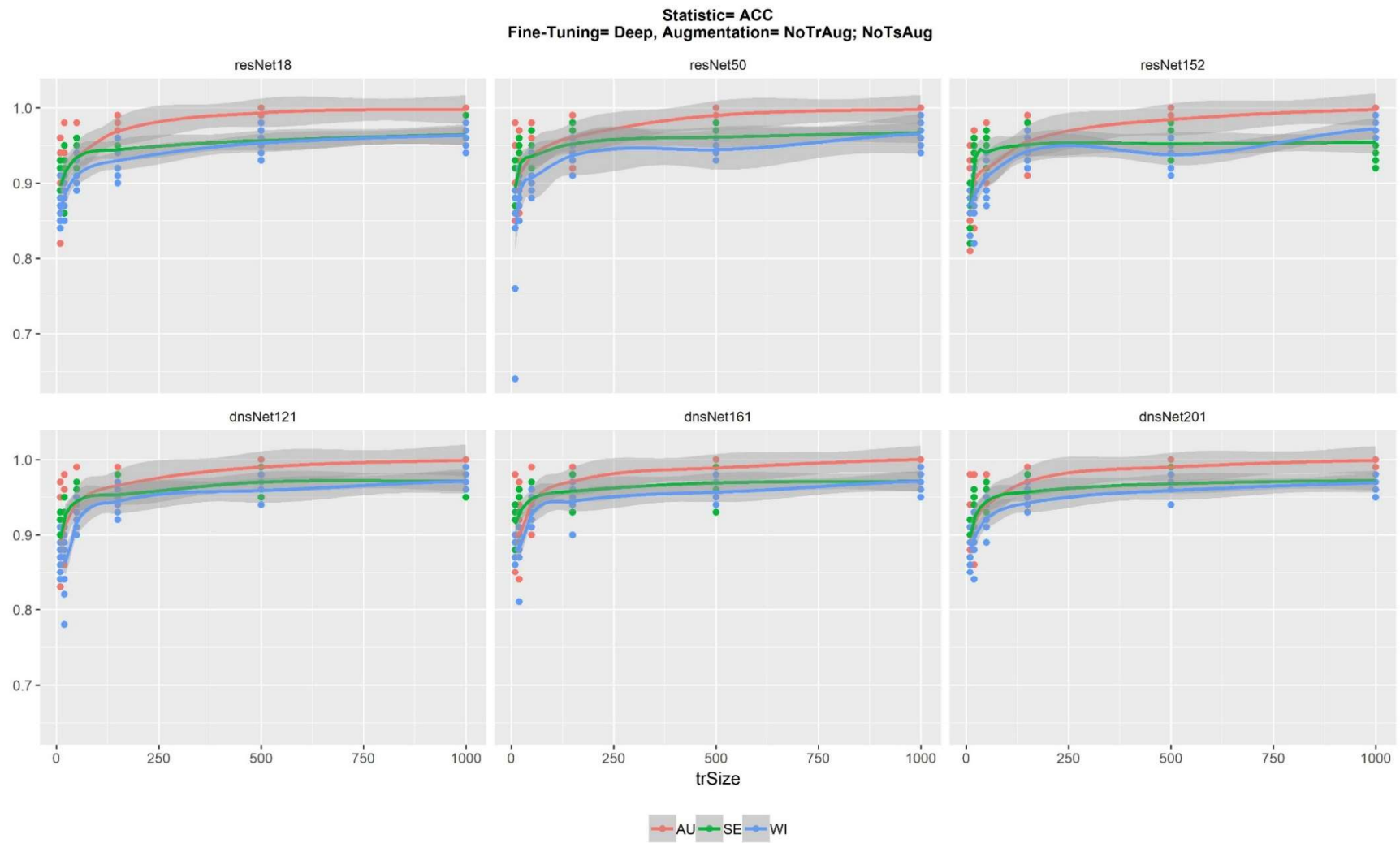

Figure A1. Overall accuracy trend with number of training images by different deep learning algorithm and three dataset. Each point represents by a point on the graph (AU: Australia, SE: Serengeti, WI: Wisconsin).

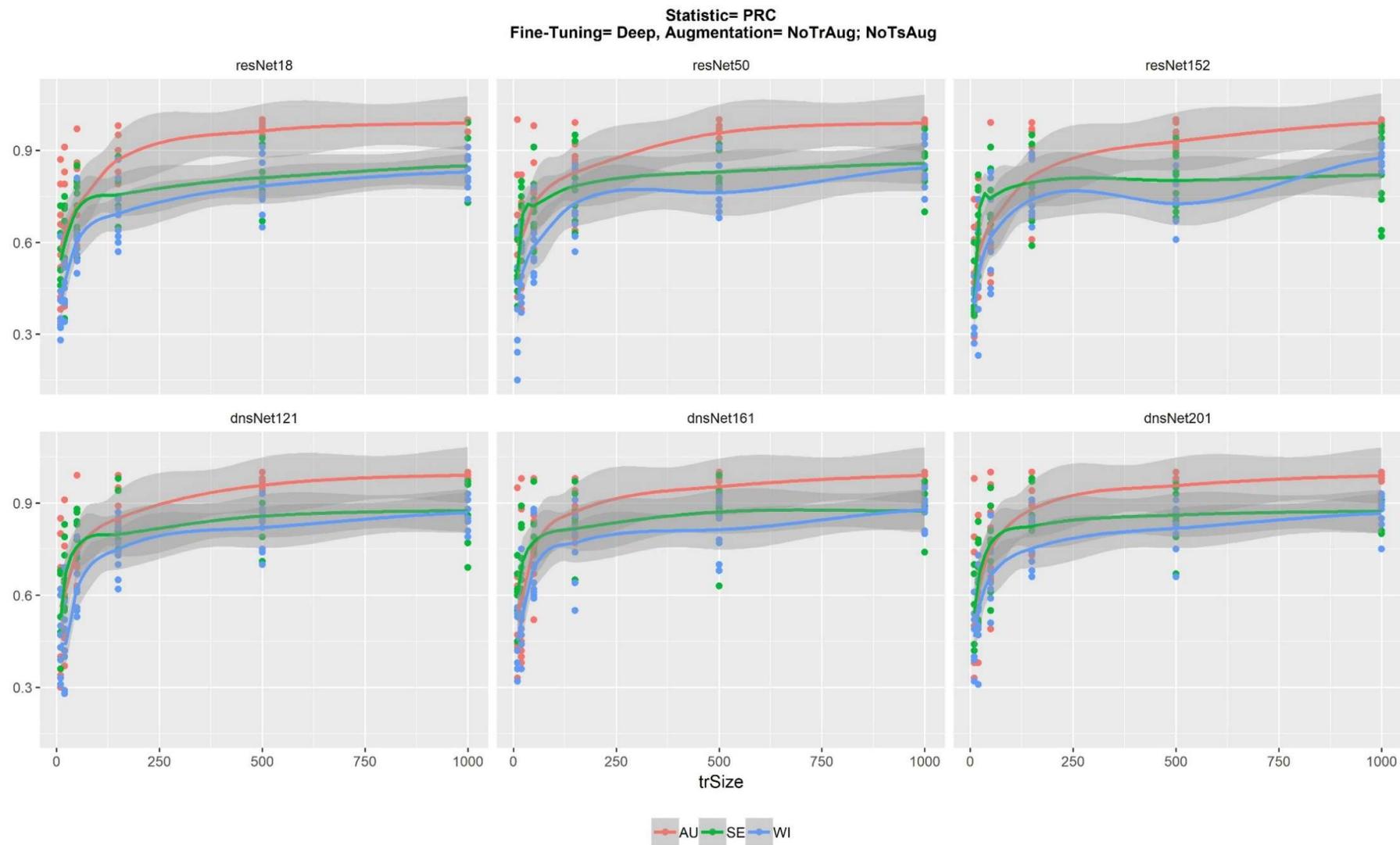

Figure A2. Precision trend with number of training images by different deep learning algorithm and three dataset. Each point represents by a point on the graph (AU: Australia, SE: Serengeti, WI: Wisconsin).

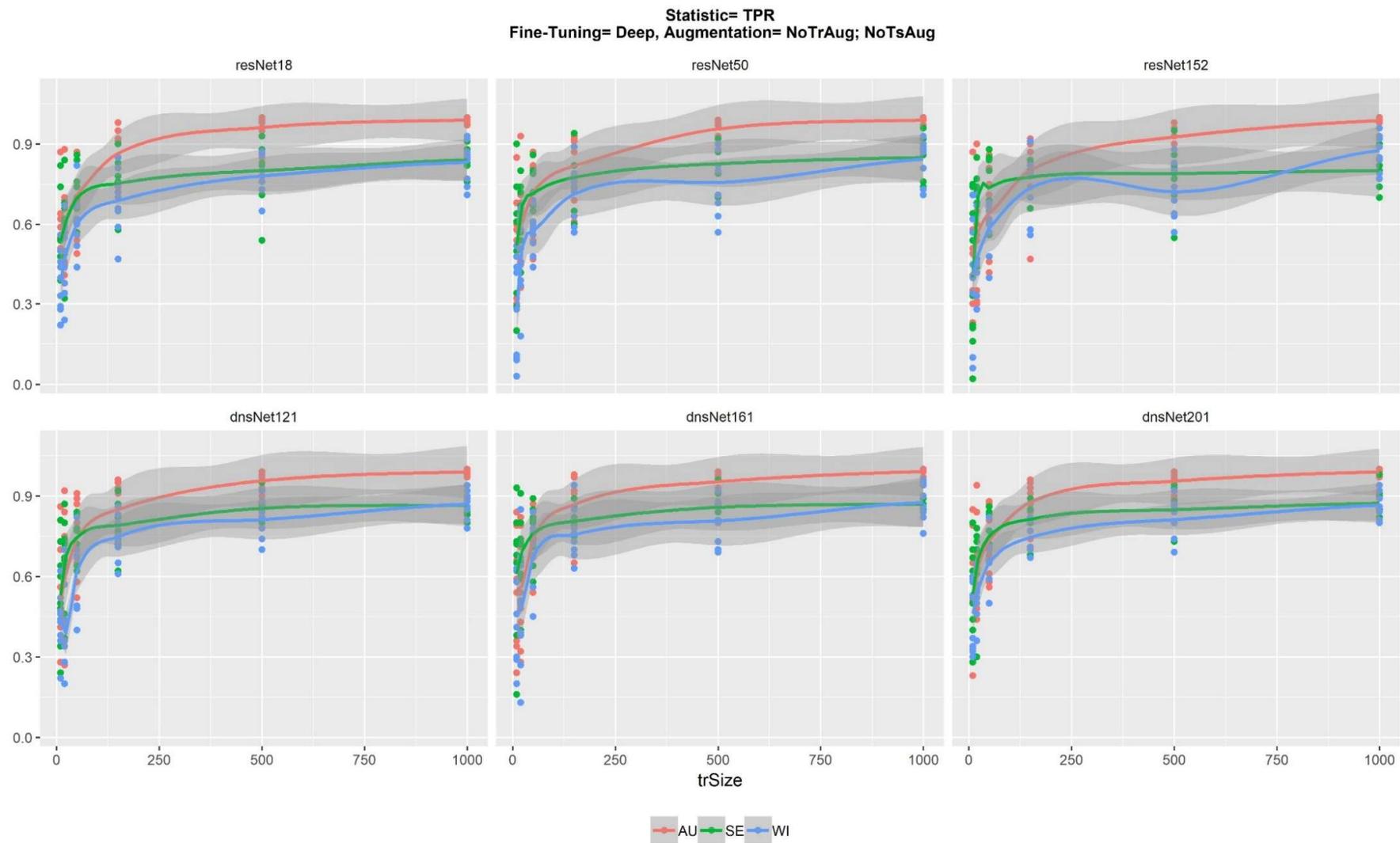

Figure A3. Recall (true positive rate) trend with number of training images by different deep learning algorithm and three dataset. Each point represents by a point on the graph (AU: Australia, SE: Serengeti, WI: Wisconsin).

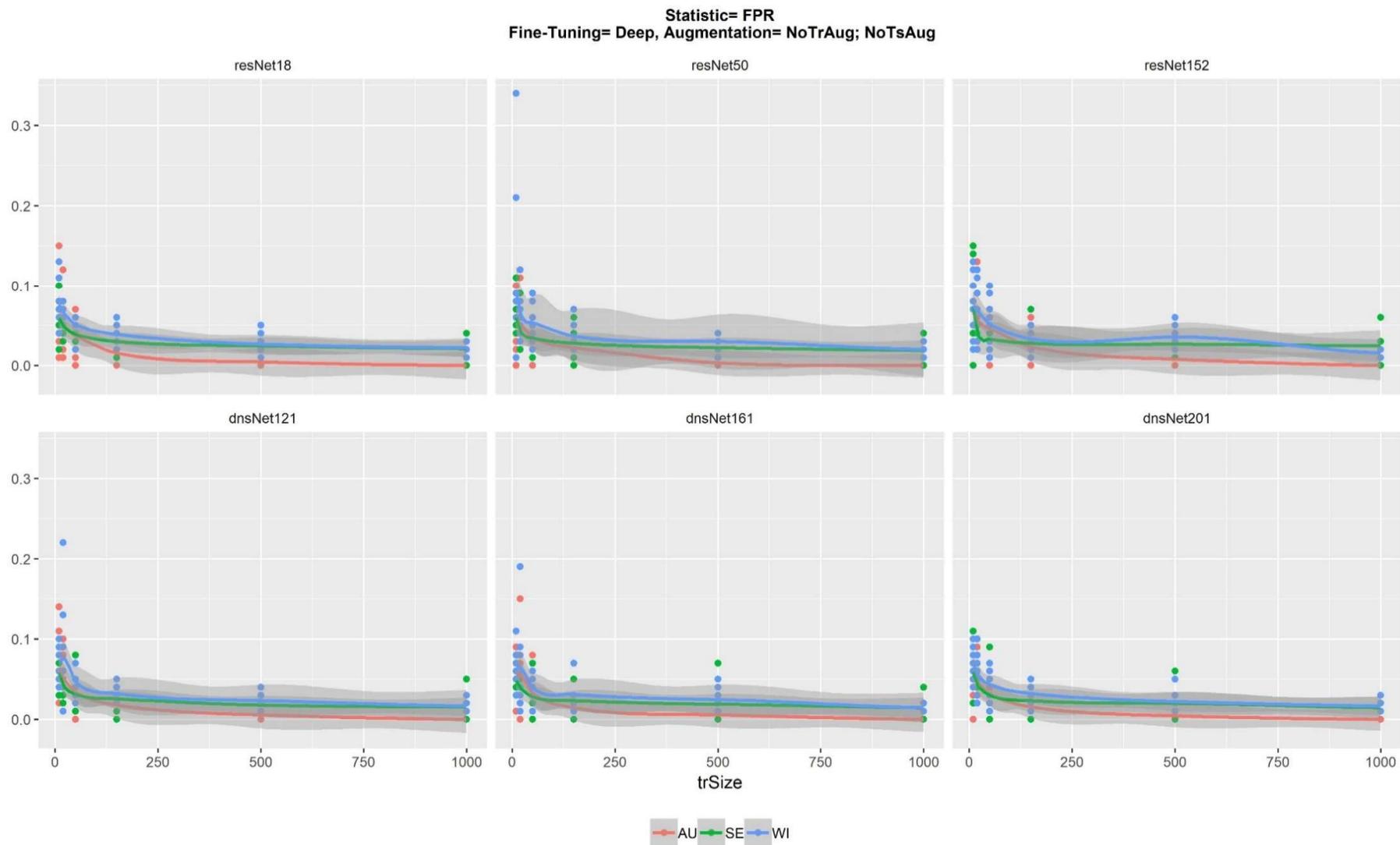

Figure A4. False positive rate trend with number of training images by different deep learning algorithm and three dataset. Each point represents by a point on the graph (AU: Australia, SE: Serengeti, WI: Wisconsin).

Table A1. Dataset composition for evaluation of Deep Learning model performance.

| Location | Class | Data 1 Train (Test) | Data 2 Train (Test) | Data 3 Train (Test) | Data 4 Train (Test) | Data 5 Train (Test) | Data 6 Train (Test) |
|---|---|---|---|---|---|---|---|
| Australia | Blank | 10 (250) | 20 (250) | 50 (250) | 150 (250) | 500 (250) | 1000 (250) |
| | Cat | 10 (250) | 20 (250) | 50 (250) | 150 (250) | 500 (250) | 1000 (250) |
| | Dog | 10 (250) | 20 (250) | 50 (250) | 150 (250) | 500 (250) | 1000 (250) |
| | Fox | 10 (250) | 20 (250) | 50 (250) | 150 (250) | 500 (250) | 1000 (250) |
| | Horse | 10 (250) | 20 (250) | 50 (250) | 150 (250) | 500 (250) | 1000 (250) |
| | Kangaroo | 10 (250) | 20 (250) | 50 (250) | 150 (250) | 500 (250) | 1000 (250) |
| | Lyrebird | 10 (250) | 20 (250) | 50 (250) | 150 (250) | 500 (250) | 1000 (250) |
| | Other | 10 (250) | 20 (250) | 50 (250) | 150 (250) | 500 (250) | 1000 (250) |
| Serengeti | Baboon | 10 (250) | 20 (250) | 50 (250) | 150 (250) | 500 (250) | 1000 (250) |
| | Blank | 10 (250) | 20 (250) | 50 (250) | 150 (250) | 500 (250) | 1000 (250) |
| | Buffalo | 10 (250) | 20 (250) | 50 (250) | 150 (250) | 500 (250) | 1000 (250) |
| | Cheetah | 10 (250) | 20 (250) | 50 (250) | 150 (250) | 500 (250) | 1000 (250) |
| | Elephant | 10 (250) | 20 (250) | 50 (250) | 150 (250) | 500 (250) | 1000 (250) |
| | Hippopotamus | 10 (250) | 20 (250) | 50 (250) | 150 (250) | 500 (250) | 1000 (250) |
| | Impala | 10 (250) | 20 (250) | 50 (250) | 150 (250) | 500 (250) | 1000 (250) |
| | Other | 10 (250) | 20 (250) | 50 (250) | 150 (250) | 500 (250) | 1000 (250) |
| Wisconsin | Bear | 10 (250) | 20 (250) | 50 (250) | 150 (250) | 500 (250) | 1000 (250) |
| | Blank | 10 (250) | 20 (250) | 50 (250) | 150 (250) | 500 (250) | 1000 (250) |
| | Elk | 10 (250) | 20 (250) | 50 (250) | 150 (250) | 500 (250) | 1000 (250) |
| | Opossum | 10 (250) | 20 (250) | 50 (250) | 150 (250) | 500 (250) | 1000 (250) |
| | Other | 10 (250) | 20 (250) | 50 (250) | 150 (250) | 500 (250) | 1000 (250) |
| | Porcupine | 10 (250) | 20 (250) | 50 (250) | 150 (250) | 500 (250) | 1000 (250) |
| | Raccoon | 10 (250) | 20 (250) | 50 (250) | 150 (250) | 500 (250) | 1000 (250) |
| | Snowshoe Hare | 10 (250) | 20 (250) | 50 (250) | 150 (250) | 500 (250) | 1000 (250) |

Table A2. Model Performance Metrics on test set (ACC: Accuracy, PRC: Precision, TPR: True Positive Rate, FPR: False Positive Rate) for the Snapshot Australia dataset. Metrics presented are stratified across class and number of training images but aggregated across model architecture, tuning and augmentation strategy (ACC ≥ 0.95; PRC ≥ 0.85; TPR ≥ 0.85; FPR ≥ 0.02 are in bold).

| MPM | NUMTRIMAGES | CLASSES | | | | | | | | | |
|---|---|---|---|---|---|---|---|---|---|---|---|
| | | BLANK | CAT | DOG | FOX | HORSE | KANGAROO | LYREBIRD | OTHERS | PIG | AVERAGE |
| ACC | 10 | **0.96 (0.019)** | 0.86 (0.022) | 0.87 (0.031) | 0.87 (0.023) | 0.93 (0.016) | 0.91 (0.019) | 0.92 (0.023) | 0.85 (0.035) | 0.86 (0.028) | 0.89 (0.045) |
| | 20 | **0.97 (0.010)** | 0.87 (0.016) | 0.91 (0.022) | 0.89 (0.016) | 0.94 (0.019) | 0.93 (0.017) | 0.94 (0.017) | 0.86 (0.025) | 0.89 (0.016) | 0.91 (0.038) |
| | 50 | **0.98 (0.004)** | 0.90 (0.013) | 0.94 (0.018) | 0.91 (0.012) | **0.96 (0.034)** | **0.95 (0.013)** | **0.96 (0.010)** | 0.90 (0.017) | 0.91 (0.014) | 0.94 (0.032) |
| | 150 | **0.99 (0.002)** | 0.93 (0.013) | **0.97 (0.008)** | **0.95 (0.016)** | **0.98 (0.005)** | **0.98 (0.007)** | **0.98 (0.006)** | 0.94 (0.015) | **0.96 (0.011)** | **0.96 (0.022)** |
| | 500 | **1.00 (0.002)** | **0.97 (0.013)** | **0.99 (0.003)** | **0.98 (0.009)** | **0.99 (0.005)** | **0.99 (0.004)** | **1.00 (0.005)** | **0.98 (0.012)** | **0.99 (0.007)** | **0.99 (0.013)** |
| | 1000 | **1.00 (0.000)** | **0.98 (0.010)** | **1.00 (0.005)** | **0.99 (0.01)** | **1.00 (0.002)** | **1.00 (0.005)** | **1.00 (0.002)** | **0.99 (0.008)** | **1.00 (0.006)** | **0.99 (0.008)** |
| PRC | 10 | **0.85 (0.125)** | 0.39 (0.054) | 0.46 (0.089) | 0.45 (0.104) | 0.71 (0.096) | 0.62 (0.103) | 0.67 (0.108) | 0.33 (0.056) | 0.38 (0.103) | 0.54 (0.194) |
| | 20 | **0.85 (0.069)** | 0.44 (0.044) | 0.62 (0.104) | 0.51 (0.096) | 0.76 (0.112) | 0.69 (0.093) | 0.73 (0.096) | 0.40 (0.076) | 0.50 (0.081) | 0.61 (0.172) |
| | 50 | **0.98 (0.017)** | 0.57 (0.064) | 0.70 (0.077) | 0.65 (0.081) | 0.79 (0.107) | 0.8 (0.071) | 0.84 (0.070) | 0.55 (0.085) | 0.62 (0.076) | 0.72 (0.153) |
| | 150 | **0.98 (0.021)** | 0.70 (0.067) | 0.83 (0.051) | 0.80 (0.087) | **0.92 (0.038)** | **0.88 (0.048)** | **0.93 (0.037)** | 0.76 (0.089) | 0.80 (0.069) | 0.84 (0.103) |
| | 500 | **1.00 (0.005)** | **0.85 (0.071)** | **0.93 (0.022)** | **0.93 (0.037)** | **0.97 (0.016)** | **0.95 (0.018)** | **0.98 (0.020)** | **0.91 (0.068)** | **0.94 (0.033)** | **0.94 (0.056)** |
| | 1000 | **1.00 (0.000)** | **0.91 (0.066)** | **0.97 (0.027)** | **0.97 (0.029)** | **0.99 (0.009)** | **0.99 (0.018)** | **0.99 (0.008)** | **0.96 (0.033)** | **0.98 (0.021)** | **0.97 (0.039)** |
| TPR | 10 | **0.87 (0.026)** | 0.47 (0.106) | 0.47 (0.152) | 0.36 (0.092) | 0.60 (0.148) | 0.59 (0.079) | 0.70 (0.119) | 0.28 (0.105) | 0.35 (0.105) | 0.52 (0.208) |
| | 20 | **0.90 (0.045)** | 0.48 (0.113) | 0.64 (0.099) | 0.42 (0.095) | 0.65 (0.094) | 0.68 (0.072) | 0.77 (0.076) | 0.39 (0.089) | 0.46 (0.133) | 0.60 (0.189) |
| | 50 | **0.87 (0.016)** | 0.56 (0.104) | 0.82 (0.067) | 0.55 (0.084) | **0.87 (0.038)** | 0.77 (0.067) | 0.80 (0.072) | 0.52 (0.065) | 0.62 (0.097) | 0.71 (0.153) |
| | 150 | **0.93 (0.021)** | 0.69 (0.076) | **0.91 (0.063)** | 0.73 (0.071) | **0.92 (0.026)** | **0.91 (0.029)** | **0.92 (0.043)** | 0.73 (0.068) | 0.83 (0.062) | 0.84 (0.108) |
| | 500 | **0.98 (0.011)** | **0.85 (0.055)** | **0.98 (0.016)** | **0.89 (0.055)** | **0.98 (0.012)** | **0.97 (0.020)** | **0.97 (0.016)** | **0.89 (0.036)** | **0.94 (0.049)** | **0.94 (0.059)** |
| | 1000 | **0.99 (0.009)** | **0.94 (0.047)** | **1.00 (0.01)** | **0.95 (0.048)** | **0.99 (0.009)** | **0.97 (0.021)** | **1.00 (0.009)** | **0.94 (0.040)** | **0.98 (0.023)** | **0.97 (0.036)** |
| FPR | 10 | **0.02 (0.023)** | 0.10 (0.035) | 0.08 (0.043) | 0.06 (0.031) | 0.03 (0.018) | 0.05 (0.025) | 0.05 (0.026) | 0.08 (0.051) | 0.08 (0.038) | 0.06 (0.041) |
| | 20 | **0.02 (0.012)** | 0.08 (0.028) | 0.05 (0.031) | 0.05 (0.022) | 0.03 (0.026) | 0.04 (0.021) | 0.04 (0.021) | 0.08 (0.036) | 0.06 (0.026) | 0.05 (0.032) |
| | 50 | 0.00 (0.002) | 0.06 (0.019) | 0.05 (0.022) | 0.04 (0.015) | 0.03 (0.037) | 0.03 (0.014) | **0.02 (0.013)** | 0.06 (0.020) | 0.05 (0.017) | 0.04 (0.026) |
| | 150 | 0.00 (0.003) | 0.04 (0.012) | **0.02 (0.010)** | 0.03 (0.019) | **0.01 (0.006)** | **0.02 (0.008)** | **0.01 (0.007)** | 0.03 (0.016) | 0.03 (0.011) | **0.02 (0.015)** |
| | 500 | 0.00 (0.000) | 0.02 (0.011) | 0.01 (0.003) | 0.01 (0.005) | 0.00 (0.004) | 0.01 (0.005) | 0.00 (0.003) | 0.01 (0.013) | 0.01 (0.006) | 0.01 (0.009) |
| | 1000 | 0.00 (0.000) | 0.01 (0.011) | 0.00 (0.005) | 0.00 (0.005) | 0.00 (0.001) | 0.00 (0.002) | 0.00 (0.000) | 0.00 (0.006) | 0.00 (0.004) | 0.00 (0.006) |

Table A3. Model Performance Metrics on test set (ACC: Accuracy, PRC: Precision, TPR: True Positive Rate, FPR: False Positive Rate) for the Snapshot Serengeti dataset. Metrics presented are stratified across class and number of training images but aggregated across model architecture, tuning and augmentation strategy (ACC ≥ 0.95; PRC ≥ 0.85; TPR ≥ 0.85; FPR ≥ 0.02 are in bold).

| MPM | NUMTRIMAGES | CLASSES | | | | | | | | | |
|---|---|---|---|---|---|---|---|---|---|---|---|
| | | BABOON | BLANK | BUFFALO | CHEETAH | ELEPHANT | HIPPOPOTAMUS | IMPALA | OTHERS | ZEBRA | AVERAGE |
| ACC | 10 | 0.87 (0.019) | 0.91 (0.02) | 0.88 (0.019) | 0.88 (0.024) | 0.92 (0.023) | 0.88 (0.019) | 0.92 (0.020) | 0.92 (0.020) | 0.91 (0.024) | 0.90 (0.029) |
| | 20 | 0.90 (0.011) | 0.94 (0.015) | 0.90 (0.012) | 0.90 (0.018) | 0.93 (0.012) | 0.92 (0.016) | **0.96 (0.010)** | **0.95 (0.015)** | **0.95 (0.013)** | 0.93 (0.026) |
| | 50 | 0.91 (0.010) | **0.95 (0.011)** | 0.92 (0.015) | 0.93 (0.011) | **0.95 (0.010)** | 0.94 (0.010) | **0.95 (0.010)** | **0.97 (0.012)** | **0.96 (0.010)** | 0.94 (0.021) |
| | 150 | 0.93 (0.011) | **0.96 (0.010)** | 0.93 (0.009) | **0.95 (0.006)** | **0.96 (0.006)** | **0.95 (0.008)** | **0.97 (0.008)** | **0.98 (0.006)** | **0.97 (0.006)** | **0.95 (0.019)** |
| | 500 | **0.95 (0.009)** | **0.97 (0.006)** | 0.94 (0.009) | **0.96 (0.009)** | **0.97 (0.006)** | **0.96 (0.010)** | **0.98 (0.006)** | **0.98 (0.006)** | **0.98 (0.006)** | **0.97 (0.015)** |
| | 1000 | **0.96 (0.012)** | **0.98 (0.005)** | **0.95 (0.009)** | **0.97 (0.008)** | **0.97 (0.007)** | **0.97 (0.007)** | **0.99 (0.004)** | **0.98 (0.005)** | **0.98 (0.005)** | **0.97 (0.013)** |
| PRC | 10 | 0.38 (0.077) | 0.65 (0.118) | 0.47 (0.053) | 0.46 (0.117) | 0.66 (0.107) | 0.48 (0.114) | 0.61 (0.081) | 0.63 (0.086) | 0.59 (0.116) | 0.55 (0.137) |
| | 20 | 0.57 (0.066) | 0.73 (0.085) | 0.54 (0.043) | 0.56 (0.106) | 0.73 (0.086) | 0.63 (0.072) | 0.78 (0.065) | 0.77 (0.074) | 0.79 (0.078) | 0.68 (0.124) |
| | 50 | 0.64 (0.060) | 0.77 (0.084) | 0.60 (0.060) | 0.69 (0.061) | 0.77 (0.058) | 0.76 (0.070) | 0.76 (0.05) | **0.87 (0.074)** | **0.89 (0.067)** | 0.75 (0.111) |
| | 150 | 0.68 (0.058) | 0.80 (0.067) | 0.67 (0.050) | 0.78 (0.046) | 0.81 (0.042) | 0.78 (0.056) | **0.86 (0.055)** | **0.92 (0.050)** | **0.94 (0.046)** | 0.80 (0.103) |
| | 500 | 0.79 (0.047) | **0.90 (0.036)** | 0.70 (0.052) | **0.85 (0.038)** | 0.84 (0.035) | 0.83 (0.069) | **0.93 (0.029)** | **0.93 (0.041)** | **0.92 (0.034)** | **0.85 (0.082)** |
| | 1000 | 0.82 (0.059) | **0.92 (0.031)** | 0.73 (0.047) | **0.86 (0.048)** | **0.87 (0.042)** | **0.88 (0.036)** | **0.96 (0.028)** | **0.94 (0.020)** | **0.95 (0.024)** | **0.88 (0.078)** |
| TPR | 10 | 0.30 (0.099) | 0.46 (0.180) | 0.52 (0.108) | 0.40 (0.090) | 0.63 (0.093) | 0.41 (0.161) | 0.82 (0.071) | 0.72 (0.058) | 0.59 (0.122) | 0.54 (0.195) |
| | 20 | 0.46 (0.071) | 0.71 (0.070) | 0.66 (0.078) | 0.45 (0.095) | 0.69 (0.072) | 0.69 (0.069) | 0.85 (0.037) | 0.77 (0.067) | 0.73 (0.040) | 0.67 (0.142) |
| | 50 | 0.56 (0.096) | 0.81 (0.066) | 0.79 (0.079) | 0.63 (0.059) | 0.73 (0.060) | 0.67 (0.071) | 0.85 (0.049) | 0.83 (0.040) | 0.76 (0.052) | 0.74 (0.114) |
| | 150 | 0.67 (0.059) | 0.85 (0.048) | 0.77 (0.064) | 0.71 (0.056) | 0.80 (0.038) | 0.77 (0.052) | 0.90 (0.038) | 0.86 (0.024) | 0.81 (0.030) | 0.79 (0.083) |
| | 500 | 0.76 (0.047) | 0.87 (0.037) | **0.85 (0.051)** | 0.77 (0.084) | 0.84 (0.032) | 0.83 (0.042) | **0.94 (0.026)** | 0.89 (0.020) | **0.87 (0.030)** | **0.85 (0.068)** |
| | 1000 | 0.83 (0.038) | **0.90 (0.029)** | **0.87 (0.042)** | 0.84 (0.037) | **0.86 (0.033)** | 0.84 (0.054) | **0.94 (0.036)** | **0.90 (0.024)** | **0.89 (0.027)** | **0.88 (0.049)** |
| FPR | 10 | 0.06 (0.027) | 0.03 (0.018) | 0.08 (0.027) | 0.06 (0.030) | 0.04 (0.024) | 0.06 (0.024) | 0.07 (0.027) | 0.06 (0.025) | 0.06 (0.030) | 0.06 (0.029) |
| | 20 | 0.05 (0.018) | 0.03 (0.018) | 0.07 (0.018) | 0.05 (0.024) | 0.03 (0.018) | 0.05 (0.018) | 0.03 (0.013) | 0.03 (0.012) | 0.03 (0.012) | 0.04 (0.022) |
| | 50 | 0.04 (0.012) | 0.03 (0.016) | 0.07 (0.022) | 0.04 (0.013) | 0.03 (0.010) | 0.03 (0.012) | 0.03 (0.010) | **0.02 (0.011)** | **0.01 (0.010)** | 0.03 (0.020) |
| | 150 | 0.04 (0.015) | 0.03 (0.014) | 0.05 (0.014) | 0.03 (0.008) | **0.02 (0.008)** | 0.03 (0.011) | **0.02 (0.009)** | **0.01 (0.007)** | **0.01 (0.007)** | 0.03 (0.017) |
| | 500 | 0.03 (0.007) | **0.01 (0.006)** | 0.05 (0.014) | **0.02 (0.006)** | **0.02 (0.006)** | **0.02 (0.012)** | **0.01 (0.005)** | **0.01 (0.006)** | **0.01 (0.004)** | **0.02 (0.014)** |
| | 1000 | **0.02 (0.010)** | **0.01 (0.005)** | **0.04 (0.011)** | **0.02 (0.007)** | **0.02 (0.007)** | **0.01 (0.006)** | **0.00 (0.006)** | **0.01 (0.005)** | **0.01 (0.005)** | **0.02 (0.013)** |

Table A4. Model Performance Metrics on test set (ACC: Accuracy, PRC: Precision, TPR: True Positive Rate, FPR: False Positive Rate) for the Snapshot Wisconsin dataset. Metrics presented are stratified across class and number of training images but aggregated across model architecture, tuning and augmentation strategy (ACC ≥ 0.95; PRC ≥ 0.85; TPR ≥ 0.85; FPR ≥ 0.02 are in bold).

| MPM | NUMTRIMAGES | CLASSES | | | | | | | | | |
|---|---|---|---|---|---|---|---|---|---|---|---|
| | | BEAR | BLANK | ELK | OPOSSUM | OTHERS | PURCUPINE | RACCOON | SNOWSHOEHARE | TURKEY | AVERAGE |
| ACC | 10 | 0.86 (0.033) | 0.87 (0.055) | 0.87 (0.028) | 0.85 (0.033) | 0.83 (0.029) | 0.86 (0.029) | 0.91 (0.017) | 0.89 (0.017) | 0.88 (0.029) | 0.87 (0.038) |
| | 20 | 0.90 (0.016) | 0.91 (0.030) | 0.87 (0.027) | 0.86 (0.033) | 0.87 (0.018) | 0.87 (0.024) | 0.89 (0.033) | 0.91 (0.016) | 0.89 (0.019) | 0.88 (0.030) |
| | 50 | 0.91 (0.018) | 0.94 (0.011) | 0.92 (0.017) | 0.90 (0.013) | 0.90 (0.017) | 0.91 (0.016) | 0.93 (0.010) | 0.94 (0.016) | 0.9 (0.017) | 0.92 (0.022) |
| | 150 | 0.94 (0.010) | **0.96 (0.008)** | 0.94 (0.010) | 0.92 (0.014) | 0.93 (0.010) | 0.93 (0.010) | **0.95 (0.007)** | **0.95 (0.012)** | 0.93 (0.012) | 0.94 (0.018) |
| | 500 | **0.96 (0.011)** | **0.98 (0.007)** | **0.97 (0.008)** | 0.94 (0.010) | **0.95 (0.008)** | 0.94 (0.010) | **0.97 (0.010)** | **0.97 (0.008)** | 0.94 (0.010) | **0.96 (0.017)** |
| | 1000 | **0.97 (0.007)** | **0.98 (0.006)** | **0.98 (0.005)** | **0.96 (0.010)** | **0.97 (0.008)** | **0.96 (0.008)** | **0.98 (0.006)** | **0.97 (0.007)** | **0.95 (0.008)** | **0.97 (0.013)** |
| PRC | 10 | 0.38 (0.099) | 0.45 (0.083) | 0.45 (0.079) | 0.33 (0.105) | 0.32 (0.054) | 0.38 (0.107) | 0.65 (0.112) | 0.50 (0.088) | 0.51 (0.112) | 0.44 (0.136) |
| | 20 | 0.61 (0.125) | 0.60 (0.129) | 0.45 (0.068) | 0.35 (0.101) | 0.44 (0.070) | 0.43 (0.092) | 0.53 (0.130) | 0.59 (0.089) | 0.51 (0.092) | 0.50 (0.132) |
| | 50 | 0.64 (0.087) | 0.82 (0.076) | 0.64 (0.075) | 0.54 (0.074) | 0.57 (0.089) | 0.60 (0.083) | 0.69 (0.067) | 0.75 (0.088) | 0.55 (0.061) | 0.64 (0.118) |
| | 150 | 0.75 (0.068) | **0.87 (0.036)** | 0.71 (0.060) | 0.64 (0.074) | 0.69 (0.061) | 0.70 (0.071) | 0.80 (0.055) | 0.80 (0.073) | 0.66 (0.069) | 0.74 (0.096) |
| | 500 | 0.81 (0.055) | **0.92 (0.038)** | **0.87 (0.043)** | 0.75 (0.053) | 0.78 (0.047) | 0.76 (0.060) | **0.86 (0.058)** | **0.88 (0.039)** | 0.70 (0.052) | 0.81 (0.083) |
| | 1000 | **0.88 (0.037)** | **0.94 (0.025)** | **0.90 (0.033)** | 0.83 (0.053) | 0.84 (0.048) | 0.84 (0.048) | **0.91 (0.029)** | **0.88 (0.037)** | 0.78 (0.041) | **0.87 (0.060)** |
| TPR | 10 | 0.34 (0.095) | 0.42 (0.109) | 0.55 (0.157) | 0.22 (0.082) | 0.44 (0.102) | 0.31 (0.118) | 0.51 (0.110) | 0.44 (0.121) | 0.52 (0.119) | 0.42 (0.153) |
| | 20 | 0.41 (0.104) | 0.55 (0.127) | 0.67 (0.126) | 0.22 (0.098) | 0.42 (0.065) | 0.40 (0.117) | 0.61 (0.166) | 0.52 (0.141) | 0.54 (0.124) | 0.48 (0.174) |
| | 50 | 0.59 (0.099) | 0.65 (0.065) | 0.69 (0.090) | 0.47 (0.068) | 0.59 (0.079) | 0.57 (0.064) | 0.77 (0.084) | 0.63 (0.104) | 0.68 (0.038) | 0.63 (0.113) |
| | 150 | 0.72 (0.046) | 0.80 (0.043) | 0.84 (0.056) | 0.60 (0.058) | 0.72 (0.062) | 0.63 (0.076) | 0.78 (0.046) | 0.76 (0.060) | 0.69 (0.089) | 0.73 (0.096) |
| | 500 | 0.81 (0.048) | **0.87 (0.040)** | 0.87 (0.049) | 0.67 (0.063) | 0.81 (0.054) | 0.74 (0.065) | **0.88 (0.032)** | 0.83 (0.060) | 0.79 (0.045) | 0.81 (0.082) |
| | 1000 | **0.87 (0.032)** | **0.91 (0.033)** | **0.91 (0.040)** | 0.79 (0.066) | **0.89 (0.031)** | 0.80 (0.043) | **0.92 (0.029)** | **0.89 (0.031)** | 0.80 (0.037) | **0.86 (0.065)** |
| FPR | 10 | 0.08 (0.041) | 0.07 (0.066) | 0.09 (0.040) | 0.07 (0.042) | 0.12 (0.038) | 0.07 (0.038) | 0.04 (0.021) | 0.06 (0.019) | 0.07 (0.043) | 0.07 (0.046) |
| | 20 | 0.04 (0.022) | 0.05 (0.032) | 0.11 (0.036) | 0.06 (0.044) | 0.07 (0.022) | 0.07 (0.032) | 0.07 (0.043) | 0.05 (0.019) | 0.07 (0.029) | 0.06 (0.037) |
| | 50 | 0.04 (0.021) | **0.02 (0.010)** | 0.05 (0.018) | 0.05 (0.016) | 0.06 (0.023) | 0.05 (0.019) | 0.05 (0.017) | 0.03 (0.014) | 0.07 (0.021) | 0.05 (0.023) |
| | 150 | 0.03 (0.013) | **0.02 (0.005)** | 0.04 (0.014) | 0.04 (0.018) | 0.04 (0.013) | 0.03 (0.014) | **0.02 (0.010)** | 0.03 (0.013) | 0.04 (0.014) | 0.03 (0.017) |
| | 500 | **0.02 (0.009)** | **0.01 (0.006)** | **0.02 (0.007)** | 0.03 (0.008) | 0.03 (0.008) | 0.03 (0.011) | **0.02 (0.010)** | **0.01 (0.006)** | 0.04 (0.012) | **0.02 (0.013)** |
| | 1000 | **0.02 (0.007)** | **0.01 (0.005)** | **0.01 (0.005)** | **0.02 (0.008)** | **0.02 (0.008)** | **0.02 (0.008)** | **0.01 (0.005)** | **0.02 (0.006)** | 0.03 (0.008) | **0.02 (0.009)** |

Table A5. Model Test Performance Metrics (ACC: Accuracy, PRC: Precision, TPR: True Positive Rate, FPR: False Positive Rate). Metrics presented are stratified across network architectures and number of training images but aggregated across data sets, class, tuning and augmentation strategy (ACC ≥ 0.95; PRC ≥ 0.85; TPR ≥ 0.85; FPR ≥ 0.02 are in bold).

| MPM | Architecture | NumTrImages | | | | | |
|---|---|---|---|---|---|---|---|
| | | 10 | 20 | 50 | 150 | 500 | 1000 |
| ACC | resNet18 | 0.88 (0.040) | 0.90 (0.036) | 0.93 (0.027) | **0.95 (0.022)** | **0.97 (0.021)** | **0.97 (0.018)** |
| | resNet50 | 0.88 (0.045) | 0.91 (0.038) | 0.93 (0.027) | **0.95 (0.021)** | **0.97 (0.019)** | **0.98 (0.016)** |
| | resNet152 | 0.88 (0.040) | 0.91 (0.034) | 0.93 (0.032) | **0.95 (0.023)** | **0.97 (0.020)** | **0.98 (0.017)** |
| | dnsNet121 | 0.89 (0.038) | 0.91 (0.038) | 0.93 (0.026) | **0.95 (0.020)** | **0.97 (0.018)** | **0.98 (0.016)** |
| | dnsNet161 | 0.90 (0.034) | 0.91 (0.037) | 0.94 (0.026) | **0.96 (0.020)** | **0.97 (0.019)** | **0.98 (0.015)** |
| | dnsNet201 | 0.89 (0.037) | 0.91 (0.031) | 0.94 (0.026) | **0.95 (0.024)** | **0.97 (0.017)** | **0.98 (0.014)** |
| PRC | resNet18 | 0.49 (0.162) | 0.55 (0.158) | 0.68 (0.132) | 0.77 (0.107) | **0.85 (0.102)** | **0.89 (0.085)** |
| | resNet50 | 0.48 (0.162) | 0.59 (0.174) | 0.69 (0.136) | 0.79 (0.107) | **0.87 (0.090)** | **0.91 (0.078)** |
| | resNet152 | 0.47 (0.163) | 0.6 (0.158) | 0.70 (0.146) | 0.79 (0.121) | **0.86 (0.093)** | **0.91 (0.084)** |
| | dnsNet121 | 0.53 (0.164) | 0.59 (0.161) | 0.71 (0.130) | 0.80 (0.099) | **0.88 (0.088)** | **0.91 (0.076)** |
| | dnsNet161 | 0.56 (0.160) | 0.62 (0.166) | 0.72 (0.134) | 0.81 (0.101) | **0.88 (0.087)** | **0.92 (0.069)** |
| | dnsNet201 | 0.53 (0.163) | 0.62 (0.139) | 0.73 (0.134) | 0.80 (0.12) | **0.88 (0.082)** | **0.92 (0.068)** |
| TPR | resNet18 | 0.47 (0.178) | 0.54 (0.176) | 0.66 (0.138) | 0.76 (0.108) | 0.84 (0.099) | **0.88 (0.079)** |
| | resNet50 | 0.46 (0.216) | 0.57 (0.180) | 0.68 (0.127) | 0.78 (0.109) | **0.86 (0.090)** | **0.90 (0.072)** |
| | resNet152 | 0.46 (0.220) | 0.59 (0.192) | 0.67 (0.160) | 0.78 (0.106) | **0.86 (0.095)** | **0.90 (0.072)** |
| | dnsNet121 | 0.51 (0.164) | 0.58 (0.197) | 0.70 (0.133) | 0.79 (0.106) | **0.87 (0.080)** | **0.91 (0.068)** |
| | dnsNet161 | 0.55 (0.195) | 0.60 (0.199) | 0.71 (0.129) | 0.81 (0.097) | **0.87 (0.083)** | **0.91 (0.064)** |
| | dnsNet201 | 0.51 (0.173) | 0.61 (0.160) | 0.71 (0.119) | 0.79 (0.110) | **0.88 (0.080)** | **0.91 (0.065)** |
| FPR | resNet18 | 0.07 (0.041) | 0.06 (0.032) | 0.04 (0.023) | 0.03 (0.016) | **0.02 (0.015)** | 0.01 (0.012) |
| | resNet50 | 0.07 (0.052) | 0.05 (0.034) | 0.04 (0.022) | 0.03 (0.017) | **0.02 (0.013)** | 0.01 (0.012) |
| | resNet152 | 0.07 (0.043) | 0.05 (0.029) | 0.04 (0.031) | 0.03 (0.019) | **0.02 (0.014)** | 0.01 (0.012) |
| | dnsNet121 | 0.06 (0.034) | 0.05 (0.036) | 0.04 (0.021) | 0.03 (0.015) | **0.02 (0.014)** | 0.01 (0.012) |
| | dnsNet161 | 0.06 (0.031) | 0.05 (0.036) | 0.04 (0.023) | **0.02 (0.016)** | **0.02 (0.014)** | 0.01 (0.010) |
| | dnsNet201 | 0.06 (0.033) | 0.05 (0.025) | 0.04 (0.023) | 0.03 (0.019) | **0.01 (0.013)** | 0.01 (0.010) |